\documentclass[journal]{IEEEtran}
\usepackage{times}  
\usepackage{helvet}  
\usepackage{courier}  
\usepackage{url}  
\usepackage{graphicx}  
\usepackage{epstopdf}
\usepackage{amsmath,amssymb} 
\usepackage{color}
\usepackage{cite}
\usepackage{booktabs}
\usepackage{threeparttable}
\usepackage{times}
\usepackage{xcolor}
\usepackage{subfigure}
\usepackage{slashbox}
\usepackage{float}
\usepackage{multirow}
\usepackage{multicol}
\usepackage{amsmath}
\usepackage{color}
\usepackage{soul}
\usepackage[utf8]{inputenc}
\usepackage[small]{caption}
\usepackage{epstopdf}
\usepackage{float}
\usepackage[colorlinks,
            linkcolor=red,
            anchorcolor=blue,
            citecolor=green,
            backref=page
            ]{hyperref}
\usepackage{cleveref}

\begin{document}

\title{Division Gets Better: Learning Brightness-Aware and Detail-Sensitive Representations for Low-Light Image Enhancement}

\author{Huake Wang, Xiaoyang Yan, Xingsong Hou \IEEEmembership{Member, IEEE}, Junhui Li, Yujie Dun, Kaibing Zhang
	\thanks{Manuscript received August -, 2021; revised -, 2022. This work was supported in part by the NSFC of China under Grant (61872286; 62272376), Key R\&D Program of Shaanxi Province of China under Grant (2020ZDLGY04-05; S2021-YF-YBSF-0094). (Huake Wang and Xiaoyang Yan contributed equally to this work.) (Corresponding author: Xingsong Hou.)}
	\thanks{Huake Wang, Xiaoyang Yan, Xingsong Hou, Junhui Li, and Yujie Dun are with the School of Information and Communications Engineering, Xi'an Jiaotong University, Xi'an 710049, China (e-mail: wanghuake@stu.xjtu.edu.cn; xyyan@stu.xjtu.edu.cn; houxs@mail.xjtu.edu.cn; mlkkljh@stu.xjtu.edu.cn;  dunyj@mail.xjtu.edu.cn). }
    \thanks{Kaibing Zhang is with the School of Computer Science, Xi'an Polytechnic University, Xi'an, 710048, China (e-mail: zhangkaibing@xpu.edu.cn). }
}

\maketitle

\begin{abstract}
Low-light image enhancement strives to improve the contrast, adjust the visibility, and restore the distortion in color and texture. Existing methods usually pay more attention to improving the visibility and contrast via increasing the lightness of low-light images, while disregarding the significance of color and texture restoration for high-quality images. Against above issue, we propose a novel luminance and chrominance dual branch network, termed LCDBNet, for low-light image enhancement, which divides low-light image enhancement into two sub-tasks, e.g., luminance adjustment and chrominance restoration. Specifically, LCDBNet is composed of two branches, namely luminance adjustment network (LAN) and chrominance restoration network (CRN). LAN takes responsibility for learning brightness-aware features leveraging long-range dependency and local attention correlation. While CRN concentrates on learning detail-sensitive features via multi-level wavelet decomposition. Finally, a fusion network is designed to blend their learned features to produce visually impressive images. Extensive experiments conducted on seven benchmark datasets validate the effectiveness of our proposed LCDBNet, and the results manifest that LCDBNet achieves superior performance in terms of multiple reference/non-reference quality evaluators compared to other state-of-the-art competitors. Our code and pretrained model will be available.
\end{abstract}


\IEEEpeerreviewmaketitle

\section{Introduction}
\IEEEPARstart{I}{mages} shot under low-light or backlit conditions are visually-terrible for viewers and also degenerate the performance of down-stream vision tasks, such as action recognition ~\cite{1,48}, object detection ~\cite{2,49}, and semantic segmentation ~\cite{3,50}. Many efforts have been tried to increase the visibility of these images to ameliorate their low quality, including upgrading imaging devices and designing image enhancement algorithms. Undoubtedly, designing effective image enhancement algorithms is a more economically feasible way.

\begin{figure*}
\centering
\includegraphics[width=0.88\textwidth]{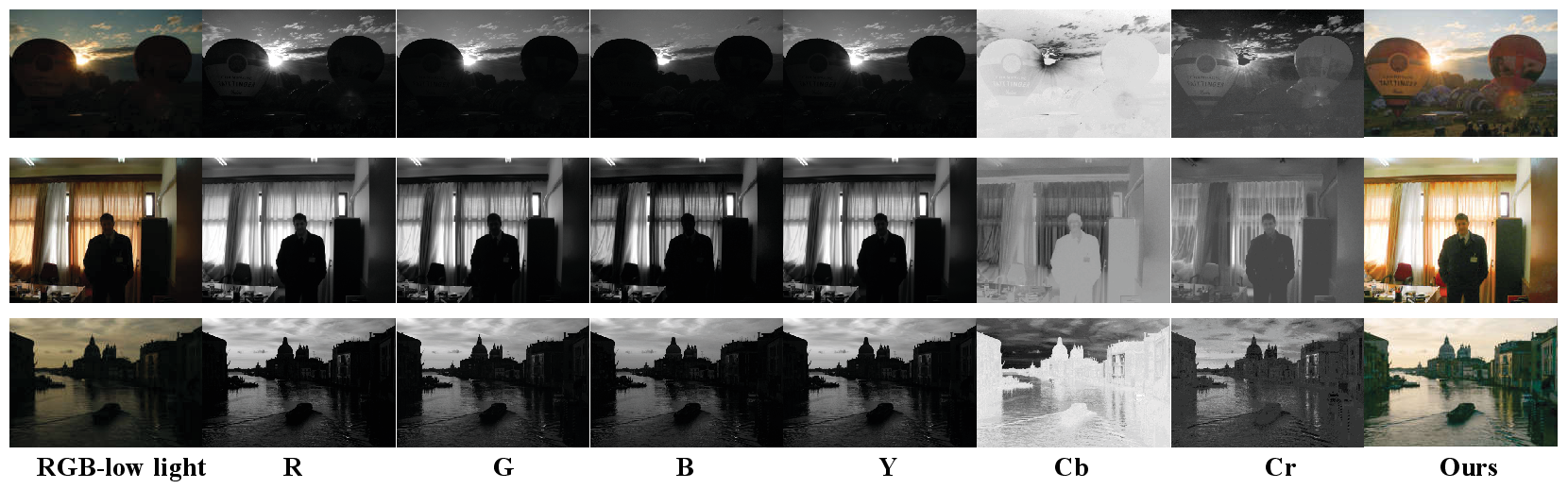}
\centering
\caption{Low-light images and their luminance and chrominance decomposition. The first four columns show original low-light images in RGB space and their RGB channel decomposition, the fifth column shows luminance (Y) channels of low-light images, the sixth and seventh columns are chrominance (CbCr) components of low-light images, and the last column exhibits the enhanced images by our method. We can obviously observe that three channels of RGB space indicate indistinguishable distortion patterns, and Y and CbCr components reveal distinctly different distortion patterns.}
\end{figure*}

In the past few decades, a large group of researchers have sought to increase the visibility of low-light images through adjusting the contrast, e.g., gamma correction ~\cite{4} and histogram equalization ~\cite{5,6}. Alternatively, Retinex theory ~\cite{7,8} postulates that the low-light image can be separated out a reflectance map as the enhanced result by eliminating the illuminance map. The above methods are too naive to enrich image details and render the image brilliance. Fueled by convolutional neural network (CNN), recent years have witnessed a significant progress in low-light image enhancement task. Roughly speaking, CNN-based methods are divided into two pipelines, e.g., end-to-end mapping ~\cite{9,10,11} and deep Retinex decomposition ~\cite{12,13,14}. End-to-end mapping learns the mapping relation from low-light images to normal-light images via auto-encoder ~\cite{9} or convolution blocks ~\cite{10,11,63}. While deep Retinex decomposition derives an enhanced image via estimating an illuminance layer or restoring a reflectance layer leveraging deep models ~\cite{15,16}. However, these methods usually operate on RGB space, which results in that designed models are hard to simultaneously learn the brightness features and detail features (including texture and color) via a single CNN.

To overcome aforementioned flaws, some methods ~\cite{17,18} developed to use the color histogram as color consistency constraint to produce vivid visual results. However, it fails to represent the local color variation of natural images. Alternatively, color space transformation is a simple and viable solution. Some visual comparisons between RGB space and YCbCr space are revealed in Figure 1. We can obviously observe from low-light images that the details of color and texture are hidden into darkness in RGB space, each channel of which show similar distortion pattern. Differently, chrominance channels show rich detail information and luminance channel carries illuminance intensity in YCbCr space. CWAN ~\cite{19} and Bread ~\cite{20} have validated the superiority of space transformation for low-light image enhancement. However, they trained several sub-networks on different channels in isolation, which could suffer from local optimal situation, resulting in failure to unlock the potential of enhancing low-light images.

Considering the above drawbacks, we present a novel luminance and chrominance dual branch network (LCDBNet) for low-light image enhancement, which can be trained end-to-end. It divides the problem of low-light enhancement into two simple sub-tasks, i.e., luminance adjustment and chrominance restoration. On the one hand, non-uniform brightness in real scene is common, therefore, global illuminance-aware and local illuminance smoothness are crucial for luminance adjustment. On the other hand, color and texture restoration focuses on pixel-wise detail refinement, so the enhancement processing needs to capture high-frequency detail information. To achieve this, we design two distinct branches, namely luminance adjustment network (LAN) and chrominance restoration network (CRN), to respectively solve them. More specifically, a global and local aggregation block (GLAB) is developed as the building block of LAN, which consists of a transformer branch ~\cite{21} and a dual attention convolution block (DACB) to learn non-local representation and local information. To recover image details, wavelet transformation ~\cite{22} is introduced to assist CRN to extract high-frequency detail information. Finally, a fusion network is proposed to combine the learned representations by LAN and CRN to produce the normal-light images. Our method can obtain favorable performance on seven benchmark datasets in terms of multiple image quality evaluators compared to other state-of-the-art methods.

In brief, the main contributions of our proposed LCDBNet are summarized as three-fold:

\begin{itemize}
  \item    We analyze the advantages of luminance and chrominance spaces for low-light image enhancement. Motivated by this, we develop a novel dual branch low-light image enhancement method, which transforms intractable image enhancement problem into two easily-handle sub-tasks: luminance adjustment and chrominance restoration.
  \item    We design a luminance and chrominance dual branch network (LCDBNet) for low-light image enhancement, which is comprised of two sub-networks, luminance adjustment network (LAN) and chrominance restoration network (CRN). LAN is used to capture brightness-aware features from the luminance channel and CRN aims to extract detail-sensitive features from the chrominance channels.
  \item    We demonstrate the superiority of proposed LCDBNet via extensive experiments on seven datasets. Our LCDBNet achieves the compelling performance and yields the visually-pleased results compared to other state-of-the-art methods.
\end{itemize}

The remainder of this paper is organized as follows. In Section \uppercase\expandafter{\romannumeral2}, the related works of low-light image enhancement are reviewed. Section \uppercase\expandafter{\romannumeral3} introduces our proposed method. Experimental details and results are shown in Section \uppercase\expandafter{\romannumeral4}. Finally, we conclude the proposed LCDBNet in Section \uppercase\expandafter{\romannumeral5}.

\begin{figure*}
\centering
\includegraphics[width=0.88\textwidth]{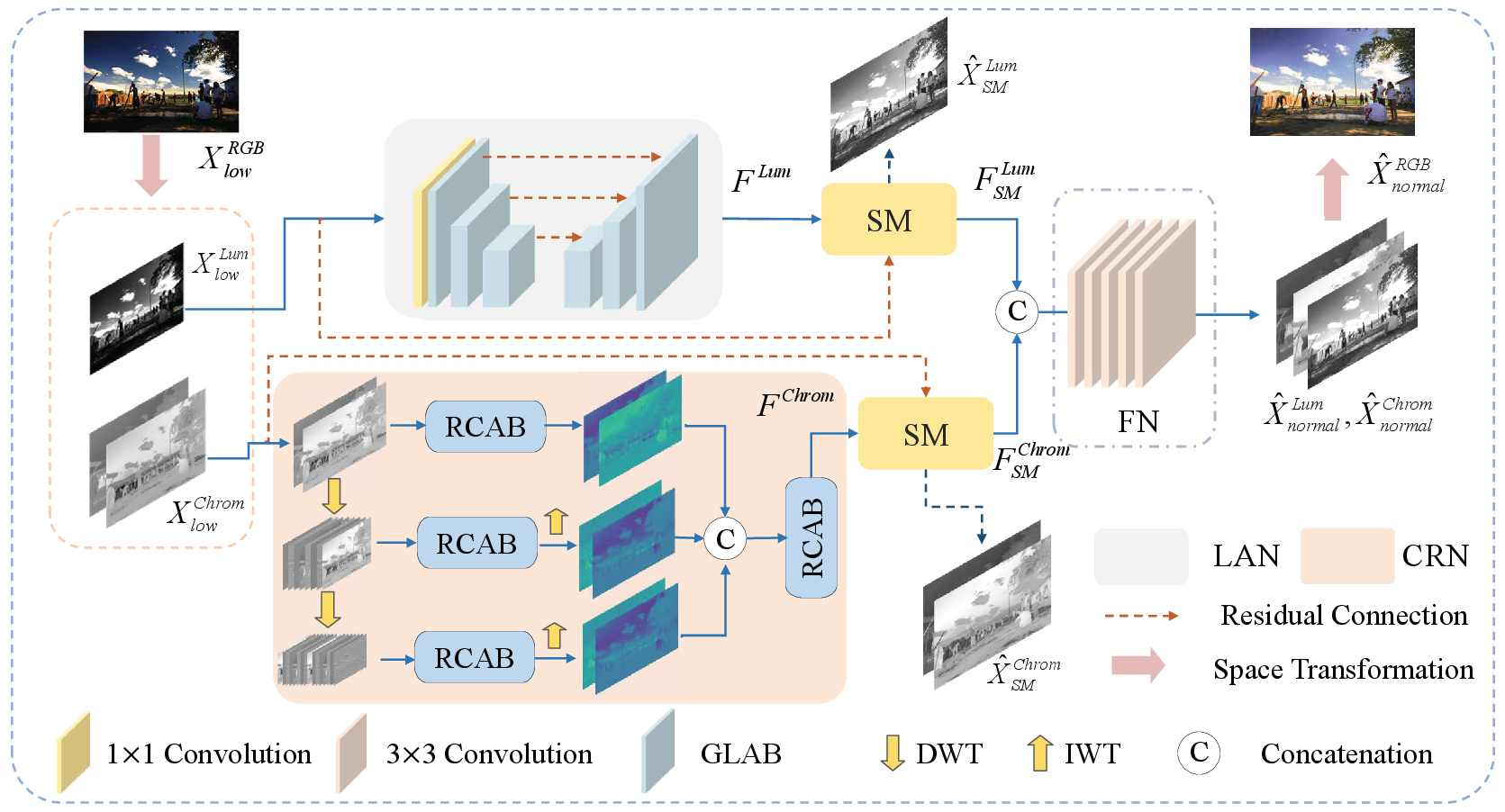}
\centering
\caption{The overview of the proposed LCDBNet. The input images are transformed from RGB space to YCbCr space. Luminance and chrominance components are fed into LAN and CRN, respectively. Then, their outputs are fused in FN to derive the enhanced results. Finally, the enhanced images are converted back to RGB space.}
\end{figure*}

\section{Related Work}
\label{sec:related}
The existing low-light image enhancement algorithms are usually divided into two categories: traditional low-light image enhancement method and CNN-based low-light image enhancement method. We will briefly retrospect these two methods in the following section. Moreover, we also review some recent transformer-based low-level vision methods.

\subsection{Traditional Method}

Histogram equalization expands the dynamic range of images to improve their contrast. CLAHE ~\cite{6} divided the global histogram into multiple local parts, which can effectively avoid the loss of details in the bright area. BPDHE ~\cite{51} used local maximum to partition histogram to preserve the brightness. To construct the relationships among neighboring pixels, LDR ~\cite{52} adopted layered difference representation of 2D histograms to amplify the image contrast. Tang et al. ~\cite{53} proposed bi-histogram equalization to handle the shifting of mean brightness. Moreover, gamma correction improves image brightness in a non-linear mapping way. BIGC ~\cite{54} used bi-coherence metric to solve the blind gamma correction. AGCWD ~\cite{4} mapped each pixel to an appropriate intensity for lightening the dark areas while suppressing the bright areas. To alleviate unwanted gamma distortion, GCME ~\cite{55} introduced a maximized differential entropy model to achieve the superior image restoration performance. 

Different from directly stretching pixel values, Retinex theory ~\cite{7, 8} decomposes images into reflectance components and illumination components, which is suitable for the enhancement of low-light images. NPE ~\cite{42} considered the trade-off between naturalness and detail by bright-pass filter and bi-log transform. SRIE ~\cite{23} employed weighted penalty items to reduce the side effect when estimating the illumination and reflectance layers. LIME ~\cite{24} constructed a structure aware constraint to quickly correct the illuminance layer via a sped-up solver. LECARM ~\cite{56} eatimated the pixel illuminance intensities based on camera response model in order to avoid color distortion. However, traditional-based methods could produce over-enhancement or unnatural results.

\subsection{CNN-Based Method}

With the increasing prosperity of deep learning, more and more researchers enhance low light images by CNNs. Lore et al. ~\cite{9} firstly proposed a deep image enhancement method, namely LLNet, which simultaneously brighten and denoise low-light images via an auto-encoder. After that, more researchers prefer to decompose the problem of low-light image enhancement into several sub-problems to reduce the difficulty of model learning and improve the performance through multi-branch networks ~\cite{10, 11, 25, 64}. For example, MBLLEN ~\cite{10} and MIRNet ~\cite{11} designed multi-branch network framework to improve the visibility of low-light images. Some methods ~\cite{18, 19} also adopted the way of divide-and-conquer, which solved several simple sub-tasks to recover image contents. DCCNet ~\cite{18} separately learned image content information and color distribution information to produce the enhanced image with vivid color and abundant detail. CWAN ~\cite{19} trained two networks to solve lightness estimation and color correction in LAB space. STANet ~\cite{57} decomposed low-light images into structure map and texture map via a contour map guided filter. Then, two different subnetworks were used to enhance them. EFINet ~\cite{58} was composed of a coefficient eatimation network for stretching pixel intensities and a fusion network for refining initial results. Besides, Wang et al. ~\cite{66} designed a flow model-based low-light enhancement method (LLFlow), which extracted illumination invariant color map to alleviate the color corruption. Zhang et al. ~\cite{68} proposed a multi-branch and progressive network, abbreviated MBPNet, to enhance the low-light image via a multi-branch framework in RGB space.

Besides, Retinex-based deep model received more attentions. Retinex-Net proposed by Wei et al. ~\cite{12} decomposed and enhanced low-light images through a DecomNet and an EnhanceNet respectively. Wang et al. ~\cite{29} proposed a progressive Retinex model to simulate ambient light and image noise through two separate networks. Moreover, KinD ~\cite{13}and KinD++ ~\cite{14} are presented to accomplish image decomposition, denoising, and enhancement via three networks. Recently, Retinexformer ~\cite{62} proposed a single-stage deep retinex model, which introduced a perturbation term to model the corruptions buried in the dark or during the enhancement process. 

\begin{figure*}
\centering
\includegraphics[width=0.88\textwidth]{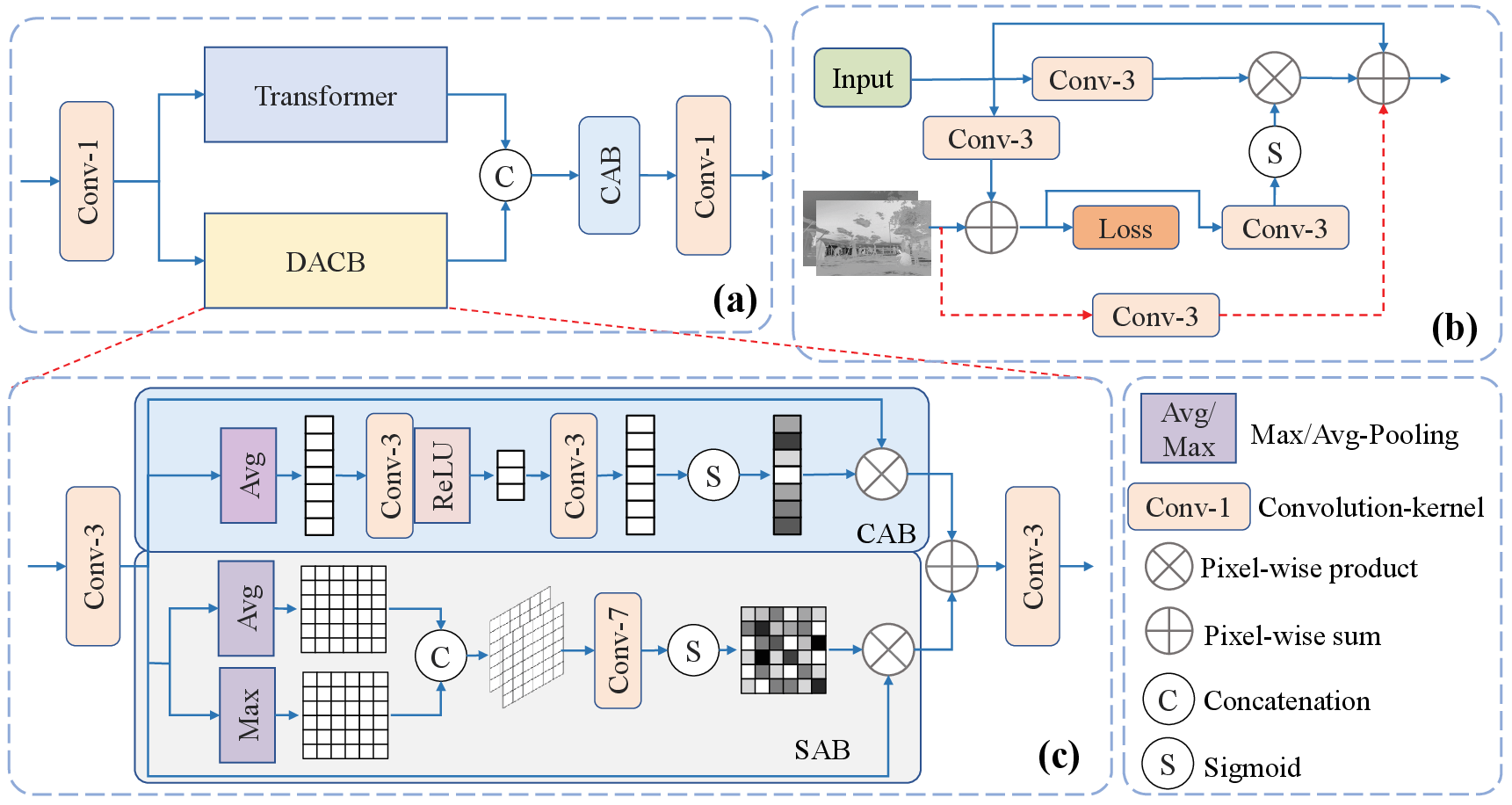}
\centering
\caption{The illustration of our designed modules. (a) is global and local aggregation block (GLAB), (b) is supervision module (SM), and (c) is double attention convolution block (DACB). CAB and SAB indicate channel attention block and spatial attention block.}
\end{figure*}

Additionally, some methods ~\cite{26,27,28}  proposed unsupervised image enhancement models to alleviate flaws of lacking paired data. Ma et al. ~\cite{67} proposed a fast low-light image enhancement framework via a self-calibrated illumination (SCI) framework without paired data. Deep image priors were also introduced to build image enhancement models without training dataset ~\cite{59,60}.

\subsection{Transformer-Based Method}
Recently, vision transformer ~\cite{30,31} has obtained the surprising results in many low-level vision tasks ~\cite{32,36}, which can extract long-range dependency to refine image content. To explore efficient transformer framework, various methods strive to reduce the computational overload via channel self-attention ~\cite{33}, local window ~\cite{34}, and multi-axis framework ~\cite{35}. Also, Wang et al. ~\cite{37} proposed a low-light transformer network (LLFormer) for low-light image enhancement, which applied an axis-based transformer block to perform self-attention on the height and width of spatial window. SNRANet ~\cite{65} exploited Transformer to restore the low-light contamination in low SNR region by means of its long-range learning ability. In short, transformer shows a promising prospect for various vision tasks. 

Different from the above methods, our proposed LCDBNet adopts a dual branch framework, performing end-to-end in YCbCr space. Considering the global illuminance variation and local smoothness, LAN combines Transformer and convolution attention to learn brightness-aware features from long-range and local perspectives. For chrominance branch, CRN aims to extract detail-sensitive representation via the information preservation ability and high-frequency decomposition ability of wavelet transform. FN aggregates brightness-aware and detail-sensitive features to produce visually-pleased enhanced images.
\section{Proposed Method}
\label{sec:method}
In this section, we first describe our motivation, and then present the overall framework of our proposed LCDBNet. Next, luminance adjustment network (LAN) and chrominance restoration network (CRN) are introduced in detail, in which we describe our design motivation for each module. Finally, we describe the framework of fusion network (FN).

\subsection{Motivation}
As shown in Figure 1, we find that each channel in RGB space of the low-light images shows similar distortions. However, luminance channel and chrominance channel reveal obvious differences when they are transformed into YCbCr space. We review the transform process from RGB space to YCbCr space as follow:
\begin{subequations}
\begin{align}
Y=&0.299 R+ 0.587 G+0.114 B,\\
Cb=&-0.147 R-0.289 G+0.436 B,\\
Cr=&0.615 R-0.515 G-0.100 B,
\end{align}
\end{subequations}
where $Y$, $Cb$, and $Cr$ denote three channels in YCbCr space and $R$, $G$, and $B$ indicate three channels in RGB space. Luminance channel $(Y)$ can be thought as a weighting combination of $R$, $G$, and $B$ channels. Hence, it contains all the information in RGB space, which is why it exhibits similar distortion to $R$, $G$, and $B$ channels. But chrominance transform can be converted to:
\begin{subequations}
\begin{align}
Cb=&0.492(B-Y),\\
Cr=&0.877(R-Y).
\end{align}
\end{subequations}

Chrominance channels refer to the weighting difference between $B (R)$ channel and $Y$ channel, which remove the interference of pixel intensity to some extent and reflect the textural structure of an image.

Above space transform inspires us to design a divide-and-conquer low-light image enhancement method, which is comprised of two branches, luminance adjustment branch and chrominance restoration branch, to process luminance and chrominance channels.

\subsection{Overall Pipeline}

The network framework of proposed LCDBNet is shown in Figure 2, which mainly involves three sub-networks, namely  LAN, CRN, and FN. LAN is used to learn brightness-aware features from luminance channel, while CRN aims to extract detail-sensitive features from chrominance channels. FN focuses on combining their derived features to produce the enhanced images. To be more concrete, given a low-light image $X^{RGB}_{low}\in \mathbb{R}^{H\times W \times 3}$, where $H\times W \times 3$ is the spatial dimension, we firstly convert it into luminance map $X^{Lum}_{low}\in \mathbb{R} ^{H\times W\times 1}$ and chrominance map $X^{Chrom}_{low} \in \mathbb{R} ^{H\times W\times 2}$ via YCbCr space transformation. Next, $X^{Lum}_{low}$ and $X^{Chrom}_{low}$ are respectively passed through LAN and CRN to extract brightness-aware features and detail-sensitive features, which are represented as:
\begin{subequations}
\begin{align}
F^{Lum}=&LAN(X^{Lum}_{low};\theta ^{LAN}),\\
F^{Chrom}=&CRN(X^{Chrom}_{low};\theta ^{CRN}),
\end{align}
\end{subequations}
where $F^{Lum}$ and $F^{Chrom}$ denotes brightness-aware features and detail-sensitive features, $\theta ^{LAN}$ and $\theta ^{CRN}$ are the parameters learned by LAN and CRN. To ensure the training efficiency and reduce the learning difficulty of LAN and CRN, two supervision modules (SM) are respectively added after the LAN and CRN, which derive the intermediate corrected luminance (chrominance) map $\hat{X}^{Lum}_{SM}$ ($\hat{X}^{Chrom}_{SM}$) and the refined brightness-aware (detail-sensitive) features $F^{Lum}_{SM}$ ($F^{Chrom}_{SM}$). Then, the refined features are  concatenated together to be fed into FN to produce the normal-light luminance map $\hat{X} ^{Lum}_{normal}$ and chrominance map $\hat{X} ^{Chrom}_{normal}$, which can be written as:
\begin{equation}
[\hat{X} ^{Lum}_{normal},\hat{X} ^{Chrom}_{normal}]=FN([F^{Lum}_{SM},F^{Chrom}_{SM}];\theta ^{FN}),
\end{equation}
where $\theta ^{FN}$ is the parameters of FN. Finally, $\hat{X} ^{Lum}_{normal}$ and $\hat{X} ^{Chrom}_{normal}$ are converted back to RGB space to derive the enhanced image $\hat{X}^{RGB}_{normal}$.

We design a novel joint loss to end-to-end train our LCDBNet, which is comprised of three sub-losses:
\begin{equation}
\mathfrak{L} =\lambda _{1}\mathbb{L}_{LAN}+ \lambda _{2}\mathbb{L}_{CRN}+\mathbb{L}_{LCDBNet},
\end{equation}
where $\mathbb{L}_{LAN}$, $\mathbb{L}_{CRN}$, and $\mathbb{L}_{LCDBNet}$ are respectively utilized to optimize LAN, CRN, and LCDBNet, and $\lambda _{1}$ and $\lambda _{2}$ are penalty parameters and are set as 0.1. Notably, our loss is operated in YCbCr space. Concretely, each sub-loss $\mathbb{L}$ contains a Charbonnier loss and an SSIM loss, which is expressed as:
\begin{subequations}
\begin{align}
\mathbb{L}=&Loss_{Char}+Loss_{SSIM},\\
Loss_{Char}=&\sqrt{ \left \| X_{normal}- \hat{X}_{normal} \right \| ^{2} + \epsilon ^{2} },\\
Loss_{SSIM}=&1-SSIM(X_{normal}, \hat{X}_{normal}),
\end{align}
\end{subequations}
where $\epsilon$ is set as 0.001 in our experiments, and $X_{normal}$ and $\hat{X}_{normal}$ are the reference image and the enhanced image. $SSIM(X, \hat{X})$ can be computed by:
\begin{equation}
SSIM(X, \hat{X}) = \frac{(2\mu_{X}\mu_{\hat{X}}+C_{1})(2\sigma_{X\hat{X}}+C_{2})}{(\mu_{X}^{2}+\mu_{\hat{X}}^{2}+C_{1})(\sigma_{X}^{2}+\sigma_{\hat{X}}^{2}+C_{2})},
\end{equation}
where $ \mu_{X}$ and $ \mu_{\hat{X}}$ are the mean of image $X$ and image $\hat{X}$, $ \sigma_{X}$ and $ \sigma_{\hat{X}}$ are the variances of image $X$ and image $\hat{X}$, $\sigma_{X\hat{X}}$ denotes the covariance of  image $X$ and image $\hat{X}$, $C_{1}$ and $C_{2}$ are two small constants to avoid zero.

\subsection{Luminance Adjustment Network}
The main purpose of LAN is to capture brightness-aware features, which can effectively correct the illuminance of low-light images. Real low-light images commonly exist global non-uniform lightness variation and local brightness smoothness, therefore, LAN is capable of modeling non-local lightness correlation and local brightness relevance. To deal with the challenges, we design a global and local aggregation block (GLAB) to capture long-range information and local spatial features, which is used as the building block of LAN. As shown in Figure 2, LAN follows the U-shape framework. It is composed of $N$ stages of encoder-decoder blocks, each stage of which consists of a GLAB and a convolution (deconvolution) layer with stride 2 for doubling (halving) the size of feature maps. Moreover, skip connection is added between corresponding encoder and decoder stages to facilitate network performance. We experimentally find that three-stage U-shape framework is sufficient for learning brightness-aware representation in comparisons to four-stage U-shape framework. Consequently, we set $N$ as 3 for LAN.

\textbf{Global and local aggregation block.} Inspired by ~\cite{36}, GLAB contains two branches, a transformer channel and a convolution channel, which respectively learn long-range dependency and local information. Due to the superiority of Swin transformer ~\cite{21}, the transformer channel only contains a Swin block. For the convolution channel, we design a double attention convolution block (DACB) to emphasize the significant local features from spatial and channel dimensions ~\cite{38}. Finally, a channel attention is used to aggregate the learned features by two branches. In a word, GLAB inherits the strength of transformer and convolution attentions, the structure of which is illustrated in Figure 3(a).

\textbf{Double attention convolution block.} Transformer block usually focuses more on the response between long distance pixels yet neglects the relations among local pixels. In fact, long-range and local information are equally important. To extract local information, DACB exploits a spatial attention branch and a channel attention branch to highlight the key feature region and channel. Then, the features learned by two branches are added together and combined via a convolution layer with kernel size of 3. The illustration of DACB is shown in Figure 3(c).

\textbf{Supervision module.} SM has two aims, one to generate the intermediate results to accelerate the training of the multi-branch network and the other to refine the learned features by previous sub-network. To achieve this, ~\cite{39} has designed such a module. However, its structure overlooks the significance of original image, which can compensate the detail information filtered out by previous convolution blocks to some extent. Hence, we add an original feature channel on the basis of ~\cite{39}, which is shown as red dashed line in Figure 3(b).

\begin{figure*}
\centering
\includegraphics[width=0.88\textwidth]{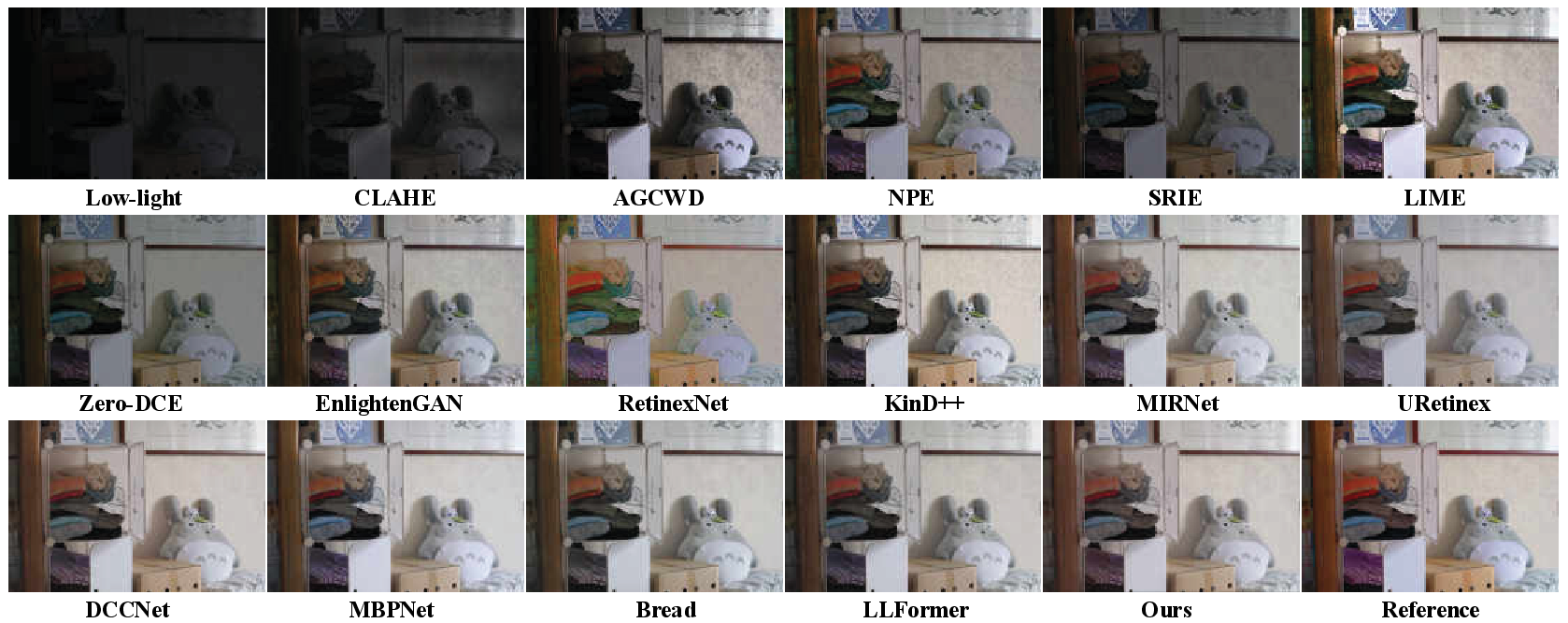}
\centering
\caption{The visual comparisons of ours and other methods on LOL dataset.}
\end{figure*}

\subsection{Chrominance Restoration Network}
CRN concentrates on capturing detail-sensitive features, which can supply some detail information, e.g., color and texture, concealed in darkness. Haar wavelet transformation can decompose an image to low-frequency layer and high-frequency layers, in which high-frequency layers convey sufficient image details that is conducive to image restoration tasks. Motivated by its advantage, CRN adopts multi-level discrete wavelet transformation (DWT) to decompose the input image to low-frequency and high-frequency sub-bands, then each-level sub-bands are passed through convolution block to learn detail-sensitive features. Since residual channel attention block (RCAB) has been proved its strength on many restoration tasks ~\cite{40}, we also use RCAB as the convolution block of CRN. Next, the learned features at each level are performed inverse wavelet transformation (IWT) to increase their resolution. Then, they are concatenated together to be fed into a RCAB to combine the multi-frequency features. Finally, the combined features are passed into SM to generate intermediate results and refined features. The overall framework of CRN is shown in Figure 2.

\subsection{Fusion Network}
FN aims to combine the learned features by previous stages to produce normal-light images. To ensure the inference efficiency, only five convolution layers with kernel size of 3 are used to aggregate the features. Then, the derived features are used to generate normal-light images via a convolution layer with output channel of 3. The detail can be found in Table I. It is necessary to emphasize that the enhanced normal-light images are in YCbCr space, which need to be reconverted back to RGB space.

\begin{table}[!t] 
  \caption{The detail of Fusion Network.}
\begin{center}
  \begin{tabular}{ccc}
  
    \toprule
    No.   &	1-5  & 6 \\
    \midrule	 
Deployment   & Conv+ReLU  & Conv\\
Kernel \& Channel   & 3\&96 & 3\&3\\			
    \bottomrule
  \end{tabular}
\end{center}

\end{table}

\section{Experimental results}
\label{sec:experimets}
In this section, dataset detail and evaluation metric are first provided, and then training detail is described. Moreover, we conduct extensive experiments on seven datasets to manifest the superiority of our proposed LCDBNet. Ablation study is performed to evaluate the effectiveness of different sub-networks. Finally, we show the running time comparisons and model complexity analysis.

\begin{table} [!t]
  \caption{The quantitative comparisons on LOL dataset in terms of PSNR, SSIM, NIQE, and LPIPS. The \textbf{bold} highlights the best results and the \underline{underline} means the second-best results.}
\begin{center}
  \begin{tabular}{ccccc}
  
    \toprule
    Methods   &	PSNR $\uparrow$	 & SSIM $\uparrow$ & 	NIQE $\downarrow$	 & LPIPS $\downarrow$  \\
    \midrule	
CLAHE	     & 8.91	      & 0.2308	 & 7.1705	 & 0.5001 \\
AGCWD	     & 13.05      & 0.4038	 & 7.8563	 & 0.4816 \\
NPE	            & 16.97	&0.5894	&9.1352	   &0.4049 \\
SRIE	     & 11.86	  & 0.4979	 & 7.5349	 & 0.3401 \\   
LIME	     & 16.76	  & 0.5644	 & 9.1272	 & 0.3945 \\ 
LECARM &14.41	&0.5688	&8.2834	&0.3262 	\\
ROPE	     & 15.02	  & 0.5092	 & 10.0985	 & 0.4713 \\  
Zero-DCE 	 & 14.86	  & 0.5849	 & 8.2230	 & 0.3352 \\ 	
EnlightenGAN & 17.48	  & 0.6578	 & 4.8878	 & 0.3223 \\ 
RetinexNet	 & 16.77	  & 0.5594	 & 9.7279	 & 0.4739 \\  
MLLEN-IC  &17.18	&0.6464	& \underline{3.2226}	&0.2253\\
KinD	     & 20.38	  & 0.8045	 & 3.9850	 & 0.1593 \\  	
KinD++	     & 21.80	  & 0.8316   & 4.0046	 & 0.1584 \\  
MIRNet	     & \underline{24.14}  & 0.8302	 &3.4786	 & 0.1311 \\ 
URetinex     & 21.33	  & 0.7906	 & 3.5431	 & \textbf{0.1210} \\
DCCNet     &22.98	     &0.7909	  &3.6716	&0.1427 \\
Bread	     & 22.96	  & 0.8121	 & 3.6826	 & 0.1597 \\
MBPNet  &22.60     &\underline{0.8332}     &3.7716 &0.1278\\
LLFormer     & 23.65	  & 0.8102	 & \textbf{3.1575}	 & 0.1692 \\   			
Ours	     & \textbf{24.21}	 & \textbf{0.8442}	 & 3.8701	 & \underline{0.1235} \\   			
    \bottomrule
  \end{tabular}
\end{center}

\end{table}

\begin{figure*}
\centering
\includegraphics[width=0.88\textwidth]{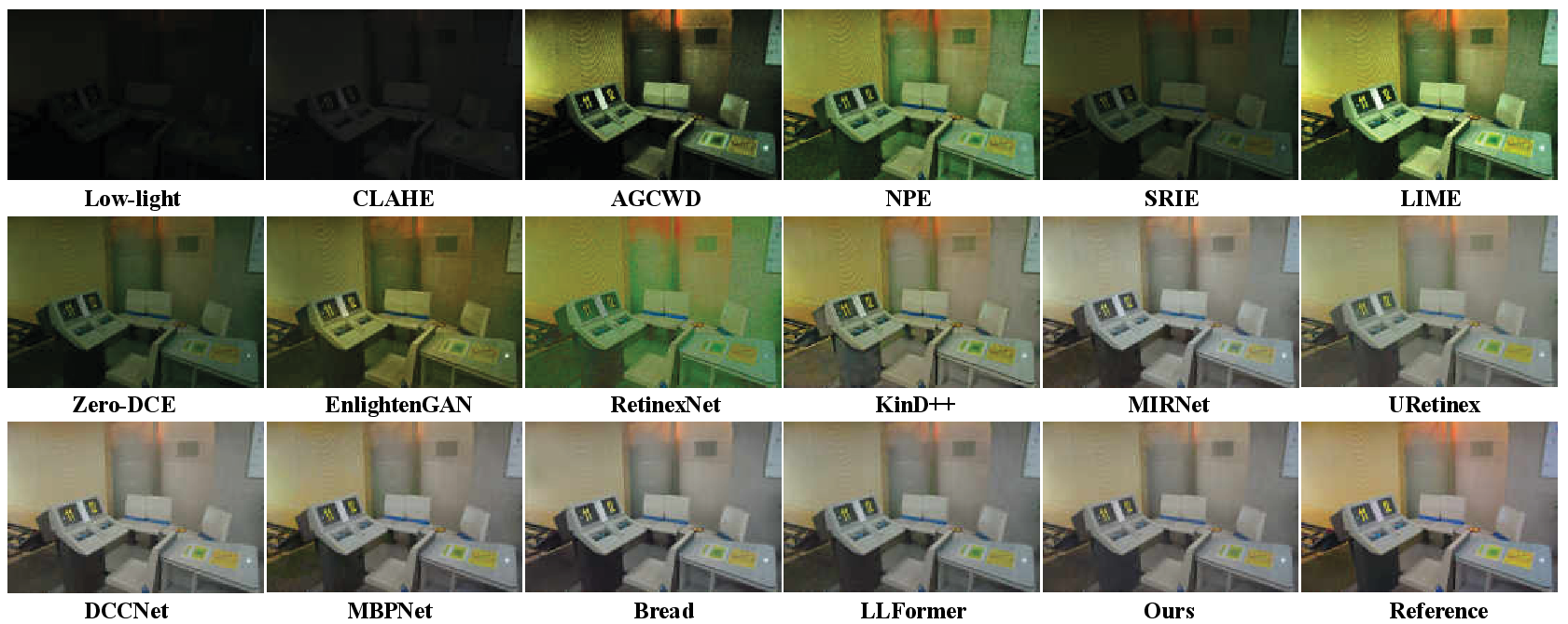}
\centering
\caption{The visual comparisons of ours and other methods on LOL dataset.}
\end{figure*}

\begin{figure*}[!t]
\centering
\includegraphics[width=0.88\textwidth]{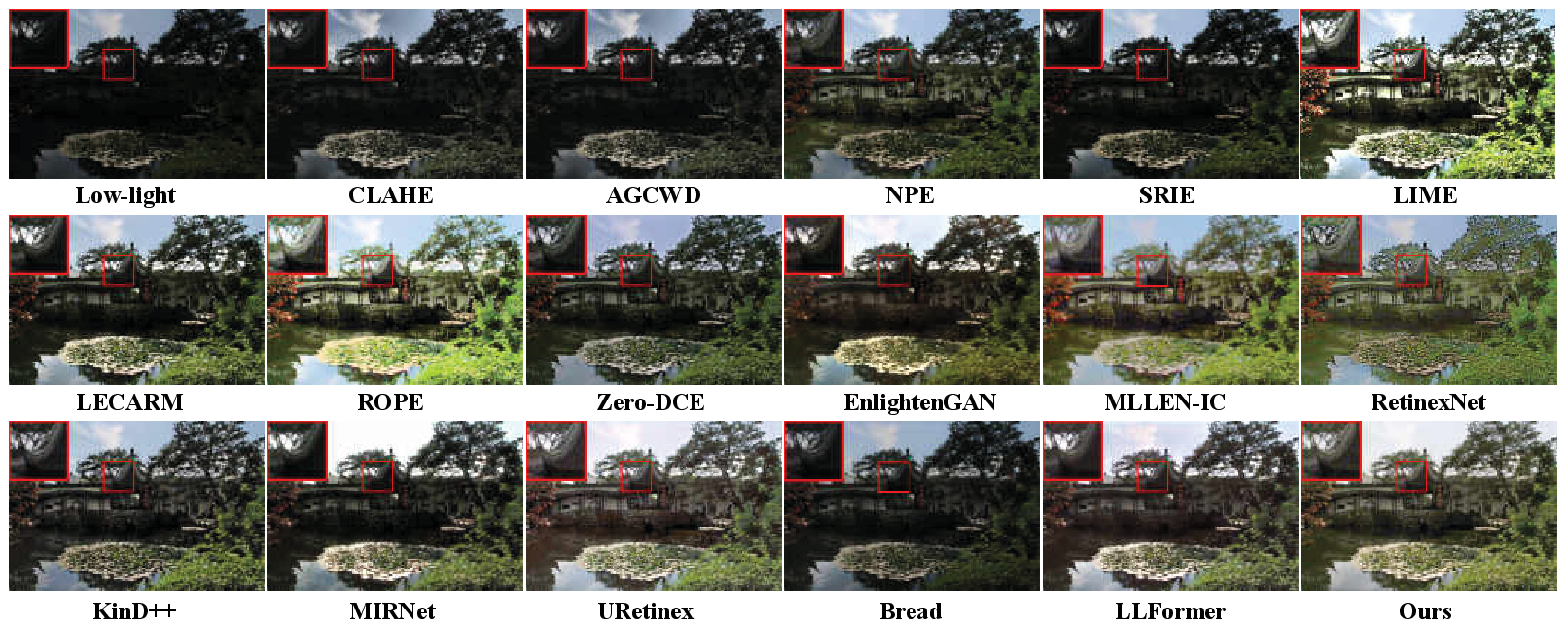}
\centering
\caption{The visual comparisons of ours and other methods on MEF dataset.}
\end{figure*}

\begin{figure*}[!t]
\centering
\includegraphics[width=0.88\textwidth]{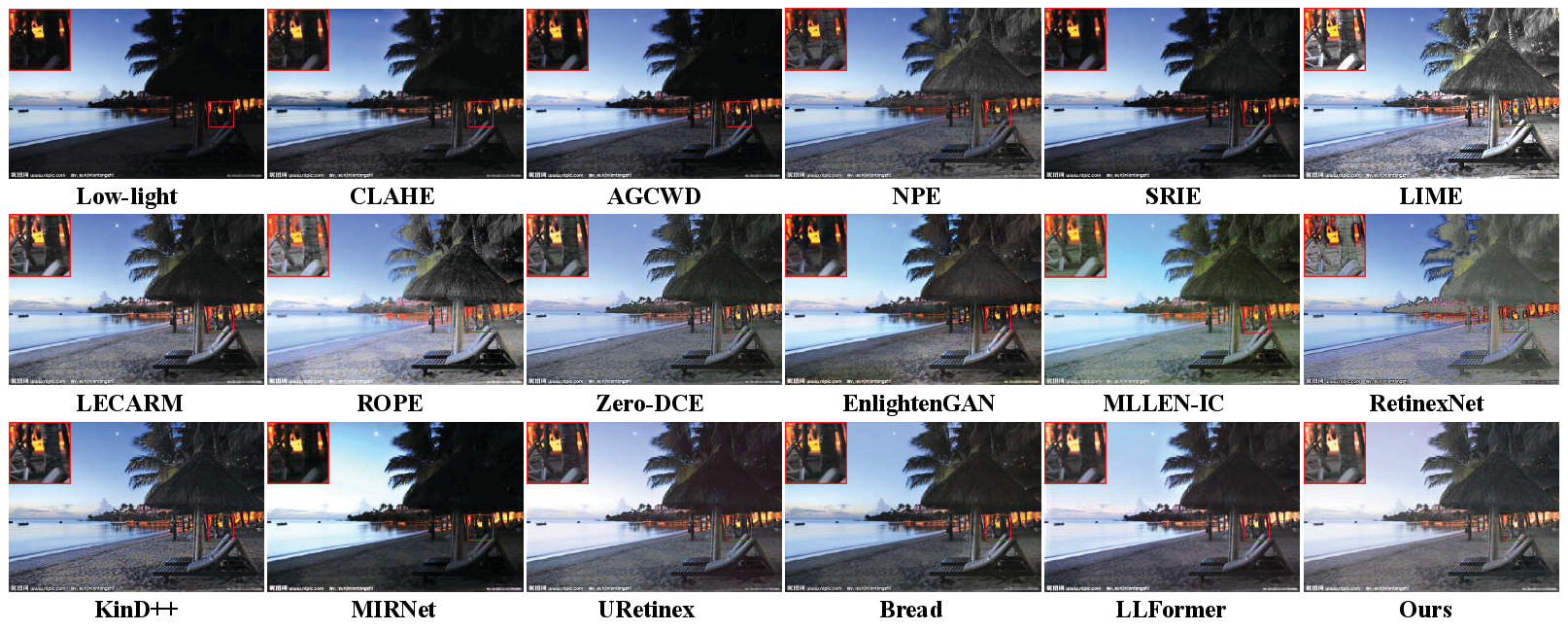}
\centering
\caption{The visual comparisons of ours and other methods on NPE dataset.}
\end{figure*}

\begin{figure*}[!t]
\centering
\includegraphics[width=0.88\textwidth]{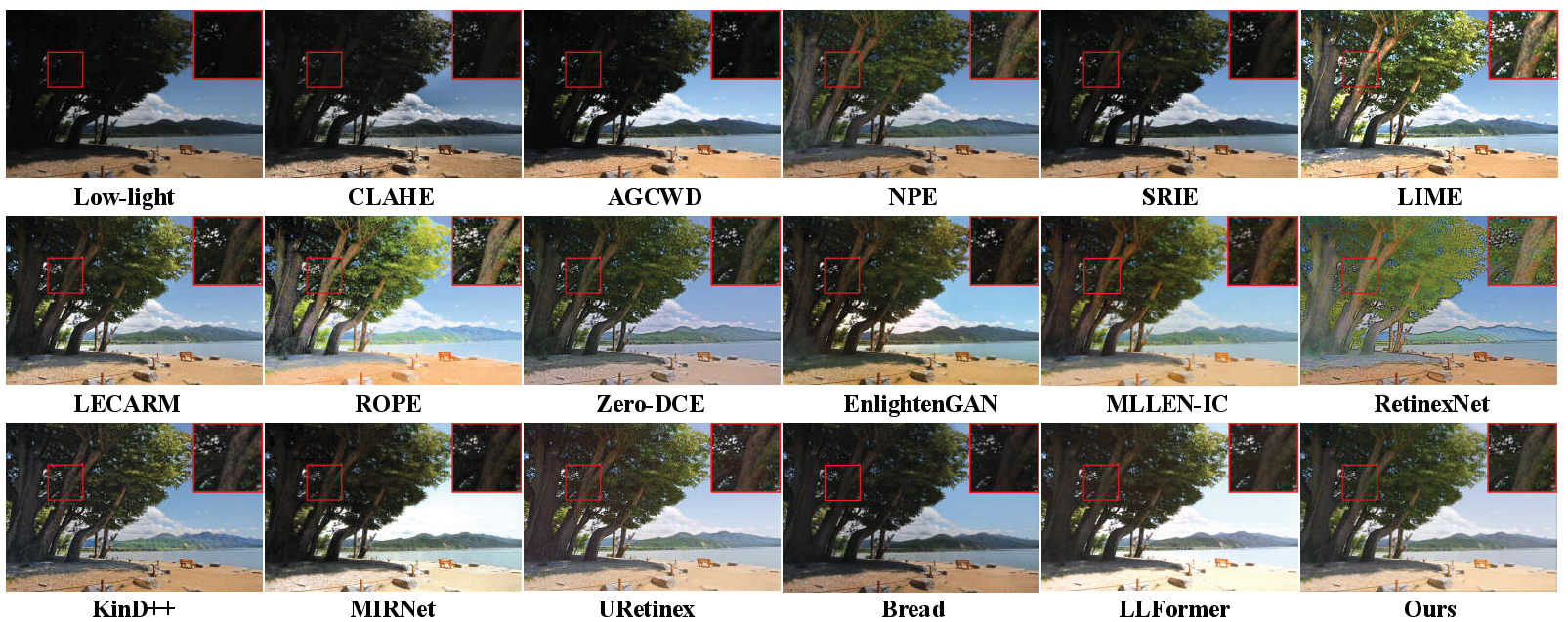}
\centering
\caption{The visual comparisons of ours and other methods on DICM dataset.}
\end{figure*}

\begin{table*} [!t]
    \caption{Comparison results on MEF, NPE, DICM, LIME, Fusion, and VV datasets in terms of NIQE. The \textbf{bold} highlights the best results and the \underline{underline} means the second-best results.}
\begin{center}
  \begin{tabular}{cccccccccc }
  
    \toprule
Methods    &Venue & Year  &MEF  & NPE  & DICM  & LIME  &Fusion  &VV  &Average \\
\midrule
CLAHE	   &GG &1994   &3.3092  &3.2647  &\underline{2.8882}  &3.6526  &3.0891  &2.3400  &3.0906\\
AGCWD	   &TIP &2013  &3.1636	&3.1810	 &2.9341  &3.6041  &3.3396  &2.3027  &3.0875\\
NPE        &TIP    &2013  & 3.5469	&3.2883	&3.0589	&3.8400	&3.2214	&2.5865	&3.2570\\
SRIE	   &CVPR &2016 &3.2041	&3.2180	 &3.3657  &\underline{3.4690}  &\underline{2.8562}  &2.1801	 &3.0322\\
LIME	   &TIP &2017  &3.8269	&3.7067	 &3.6343  &4.3473  &3.4933	&2.3276	 &3.5560\\
LECARM &TCSVT &2019 &3.1990	&3.3598	&3.7955	&3.8581	&3.2338	&2.4769	&3.3205\\
ROPE	   &ICASSP &2021&3.5657	&3.3902	 &3.2273  &4.0329  &3.5417	&2.6346	 &3.3987\\
Zero-DCE   &CVPR &2020 &3.3088	&3.5025	 &3.0973  &3.7890  &3.3948	&2.7526	 &3.3075\\
EnlightenGAN&TIP &2021&\underline{2.8923}	&3.3464	 &3.0561  &\textbf{3.3802}  &\textbf{2.7896}  &3.4542	 &3.1531\\
RetinexNet &BMVC &2018 &4.9043	&4.3896	 &4.3143  &4.9077  &3.9910	&2.6196	 &4.1878\\
MLLEN-IC  &TCSVT &2022 &3.0716	&\underline{3.0767}	&\textbf{2.8315}	&3.5674	&2.9658	&2.2338	&\underline{2.9578}\\
KinD	   &MM &2019  &3.5598	&\textbf{3.0231}	 &3.5135  &3.5825  &3.0515  &2.3991	 &3.1883\\
KinD++	   &IJCV &2021&3.3922	&3.5270	 &3.3258  &4.6313  &3.1601	&2.3710	 &3.4012\\
MIRNet	   &ECCV &2020&3.1915	&3.3391	 &3.1533  &3.5015  &3.3446	&2.6070	 &3.1895\\
URetinex   &CVPR &2022&3.2635	&3.6098	 &3.2475  &4.1808  &3.2813	&\underline{2.1726}	 &3.2926\\
Bread	   &IJCV &2023&3.5677	&3.2465	 &3.4063  &4.1323  &3.2244	&2.4788	 &3.3427\\
LLFormer   &AAAI &2023&3.2847	&3.3003	 &3.5154  &3.9295  &3.5218  &2.9112	 &3.4105\\
Ours	   & && \textbf{2.8355}	&3.1792	 &3.1369  &3.4928  &2.9625  &\textbf{2.0909}  &\textbf{2.9496}\\
    \bottomrule
  \end{tabular}
\end{center}

\end{table*}

\subsection{Dataset and Metric}
LOL dataset ~\cite{12} is a common benchmark for low-light image enhancement, which contains 485 pair of normal/low-light images for training, 15 pair of normal/low-light images for testing. In addition, many unpaired datasets are used to evaluate the performance of low-light image enhancement models, which include MEF ~\cite{41}, NPE ~\cite{42}, DICM ~\cite{43}, LIME ~\cite{24}, Fusion ~\cite{44}, and VV $\footnote{https://sites.google.com/site/vonikakis/datasets}$ datasets. Notably, we train two models on LOL training set and LOL synthetic dataset for evaluating LOL test set and these unpaired datasets. To quantitatively analyze the performance of our model, three full-reference image quality evaluators, PSNR, SSIM, and learned perceptual image patch similarity (LPIPS) ~\cite{46}, are used to evaluate the paired test set. One no-reference image quality evaluator, natural image quality evaluator (NIQE) ~\cite{47}, is used to assess unpaired datasets.

\subsection{Training Detail}

\begin{figure*}[!t]
\centering
\includegraphics[width=0.88\textwidth]{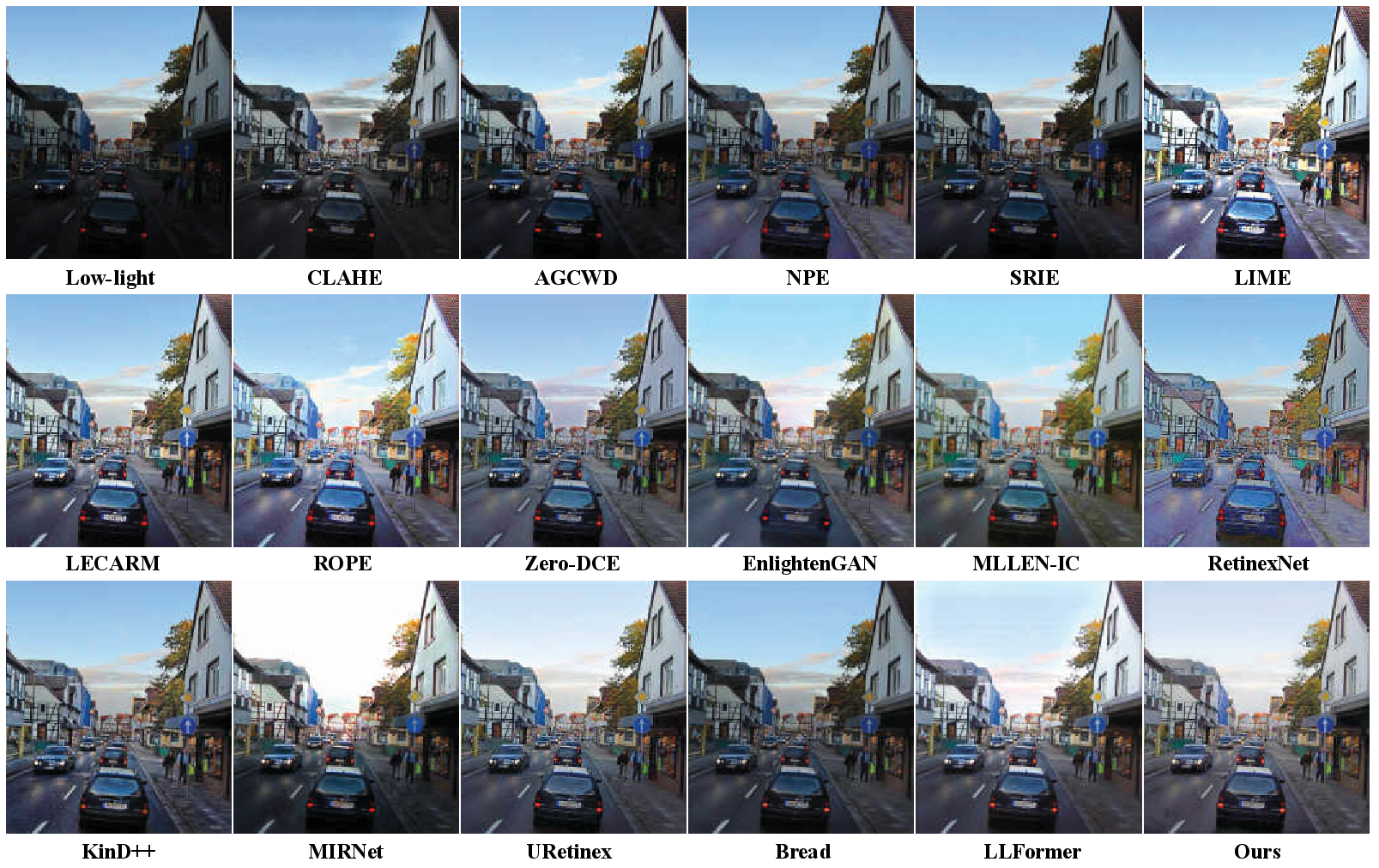}
\centering
\caption{The visual comparisons of ours and other methods on LIME dataset.}
\end{figure*}

\begin{figure*}[!t]
\centering
\includegraphics[width=0.88\textwidth]{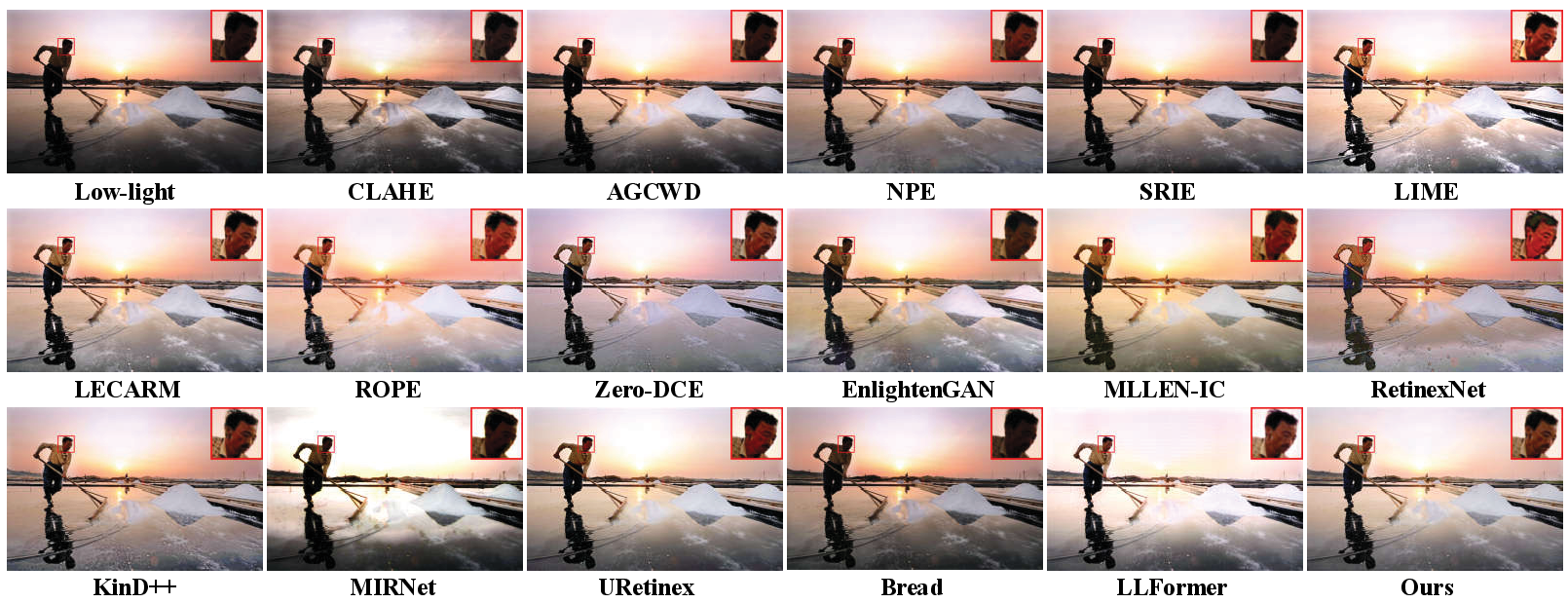}
\centering
\caption{The visual comparisons of ours and other methods on Fusion dataset.}
\end{figure*}

During training, we randomly crop the training images to a patch with spatial size of 128$\times$128, then randomly flip and rotate them for data augmentation. Our proposed LCDBNet is optimized via Adam optimizer with default parameters $\lambda_{1} = 0.9$ and $\lambda_{2} = 0.99$. Batch size and training epoch are respectively set as 8 and 2000. Moreover, the initial learning rate is given as $1\times 10^{-4}$ and attenuated to $1\times 10^{-6}$ by cosine annealing strategy. All the experiments are performed in an Inter Core i5-10400F CPU with 2.90GHz and a single GeForce RTX 2080Ti GPU computational platform. Test codes of all compared methods are downloaded from original paper.

\subsection{Evaluations and Comparisons}

We compare our proposed LCDBNet with seventeen low-light image enhancement methods, including CLAHE ~\cite{6}, AGCWD ~\cite{4}, NPE ~~\cite{42}, SRIE ~\cite{23}, LIME ~\cite{24}, LECARM ~\cite{56}, ROPE ~\cite{5}, Zero-DCE ~\cite{26}, EnlightenGAN ~\cite{28}, RetinexNet ~\cite{12}, MLLEN-IC ~\cite{61}, KinD ~\cite{13}, KinD++ ~\cite{14}, MIRNet ~\cite{11}, URetinex ~\cite{16}, DCCNet ~\cite{18}, Bread ~\cite{20}, MBPNet ~\cite{68}, and LLFormer ~\cite{37}. The comparison results on LOL dataset are reported in Table II. It is obvious to see that our method surpasses all compared low-light image enhancement models in PSNR and SSIM and obtains the second-best performance in LPIPS evaluator. For NIQE, our method performs slightly weaker than LLFormer. In brief, the objective results in Table II demonstrate the advantage of our proposed method.

In order to comprehensively compare with other methods, visual comparison results are manifested in Figure 4 and Figure 5. As can be seen, our method recovers more vivid color and richer texture. Specifically, CLAHE, AGCWD, and SRIE fail to significally enhance the lightness of low-light images. NPE and LIME leave a lot of noise in lighten areas. Zero-DCE and EnlightenGAN produce under-enhancement results. The enhancement results of RetinexNet looks unnatural and noisy. Kind++ encounters the color deviation. MIRNet and MBPNet produces some visual artifacts. The images lighten by Bread show over-smooth due to its trained denoiser. LLFormer and URetinex could produce some unpleased blur artifacts. However, our method yields visually-pleased enhanced images with abundant details. It demonstrates that our LCDBNet can learn detail-sensitive features and brightness-aware features to rejuvenate the radiance of low-light images.

\begin{figure*}[!t]
\centering
\includegraphics[width=0.88\textwidth]{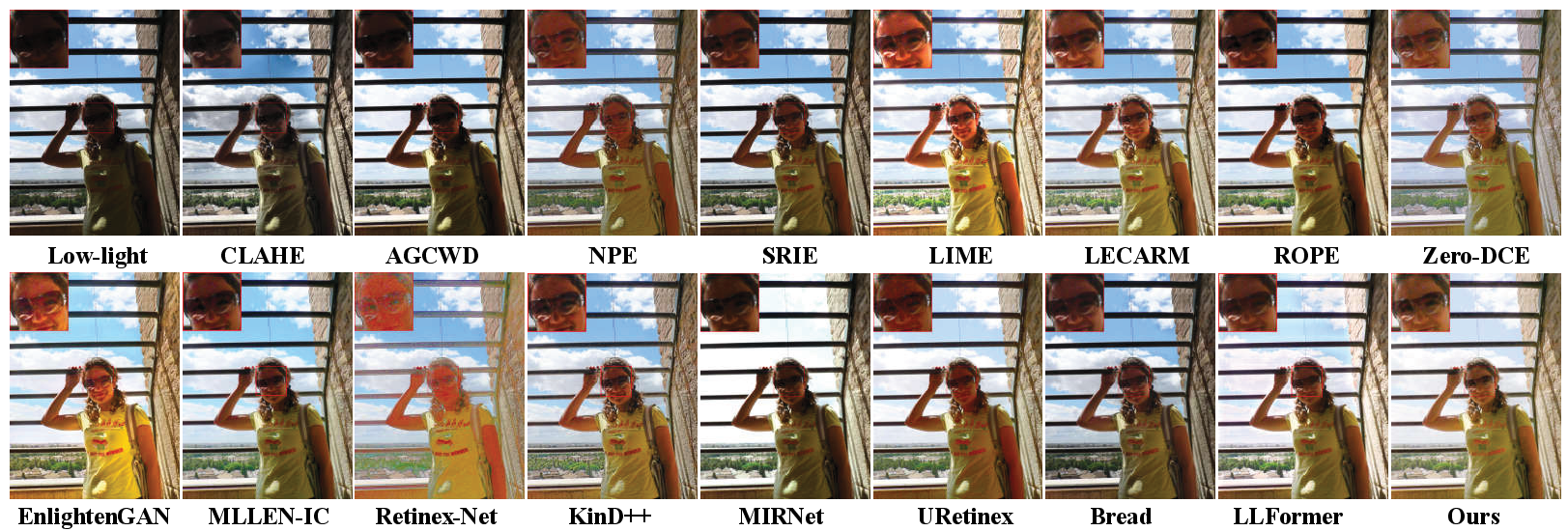}
\centering
\caption{The visual comparisons of ours and other methods on VV dataset.}
\end{figure*}

\begin{figure*}[!t]
\centering
\includegraphics[width=0.9\textwidth]{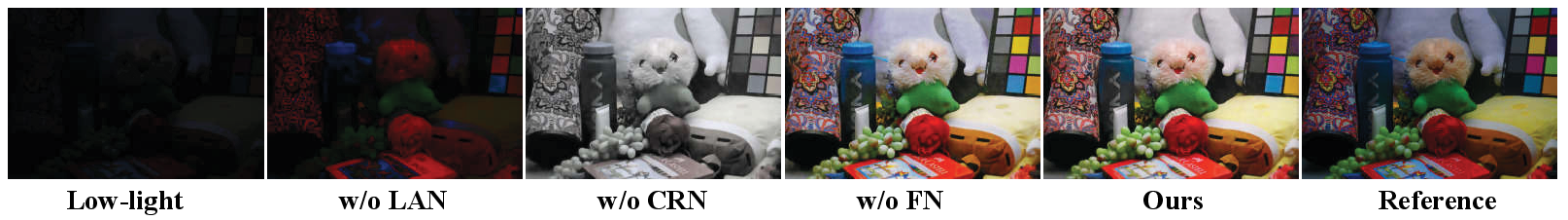}
\centering
\caption{The visual comparisons of ablation study.}
\end{figure*}

The comparison results of six unpaired test sets are tabulated in Table III. Our LCDBNet outperforms other enhancement models on MEF and VV test sets. Moreover, we show the average results among six test sets and our method achieves the best performance. It demonstrates that our LCDBNet has a robust generalization ability for various real low-light scenes. 

To validate the visual superiority of our LCDBNet, more qualitative comparisons are provided in this paper. The visual results of MEF, NPE, DICM, LIME, Fusion, and VV datasets are shown in Figure 6, Figure 7, Figure 8, Figure 9, Figure 10, and Figure 11, respectively. One can clearly observe from Figure 6 that CLAHE, AGCWD, NPE, SRIE, and LECARM cannot lighten the low-light images. LIME, ROPE, and MLLEN-IC produce over-exposure results. RetinexNet shows unnatural enhanced results. The images lightened by KinD++, MIRNet, and Bread exist local under-enhancement areas. LLFormer arises grid artifacts because of its transformer framework only learning long-range correlation of images. However, our method not only improves the brightness of low-light images, but also recovers the sharp details. It demonstrates the effctiveness of two separate sub-networks, luminance adjustness and chrominance restoration.

Figrue 7 and Figure 8 reveal similar enhancement results with Figure 6. we observe that some methods obtain good NIQE values, but their visual resluts looks worse than ours. It shows the inconsistence between non-reference evaluation metric and subjective visual result to some extent. In Figure 9, traditional low-light enhancement models generates under-enhancement or over-enhancement results. some deep learning-based enhancement methods fail to rejuvenate the radiance of images. Bread produces over-smooth results and MIRNet over-enhances the air regions. Compared to the grid artifacts of LLFormer, our result yields refine textures. It verifies the effectiveness of combination between transformer and convolution blocks in our proposed LCDBNet. In Figrue 10 and Figure 11, we zoom in the head regions of a man and a woman for a clear view. We can see that our method produces sharp facial texture and recover fascinating scene radiance.

\begin{table}[!t] 
  \caption{The ablation study for different sub-networks. The \textbf{bold} highlights the best results.}
\begin{center}
  \begin{tabular}{ccccc}
  
    \toprule
    Methods   &	w/o LAN	 & w/o CRN & w/o FN  & Ours \\
    \midrule	 
PSNR $\uparrow$   & 8.03    & 20.43   & 22.08  & \textbf{24.21}\\
SSIM $\uparrow$   & 0.2193  & 0.5794  & 0.8153 & \textbf{0.8442}\\			
    \bottomrule
  \end{tabular}
\end{center}

\end{table}

\subsection{Ablation Study}

\begin{figure}[!t]
\centering
\includegraphics[width=0.46\textwidth]{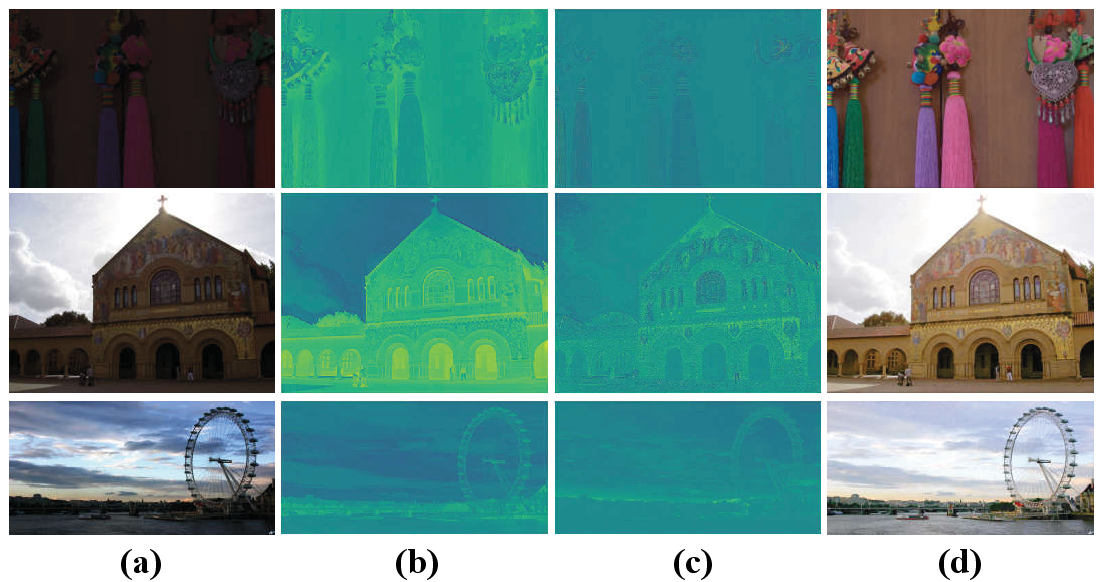}
\centering
\caption{The illustration of brightness-aware and detail-sensitive features.(a) shows low-light images, (b) reveals brightness-aware features, (c) means detail-sensitive features, and (d) indicates enhanced images}
\end{figure}

To investigate the effectiveness of different sub-networks, we perform comparison experiments on LOL dataset via removing corresponding sub-network. The results are reported in Table IV. The results of w/o LAN and w/o CRN validate the effectiveness of luminance adjustment and chrominance restoration. Moreover, the performance of w/o FN demonstrates FN can sufficiently combine brightness-aware and detail-sensitive features to produce best-optimal results. Their visual comparisons are presented in Figure 12. We can see that w/o LAN cannot lighten low-light image and w/o CRN fails to restore detail and color information. Adding FN can produce refine detail and vivid color.  In short, above experiment results prove the effectiveness of our LCDBNet.

To better show the learned features, we visualize brightness-aware and detail-sensitive features in Figure 13. It can be seen that brightness-aware features focus on darken areas while detail-sensitive features emphase the texture regions.

\begin{figure}[!t]
\centering
\includegraphics[width=0.46\textwidth]{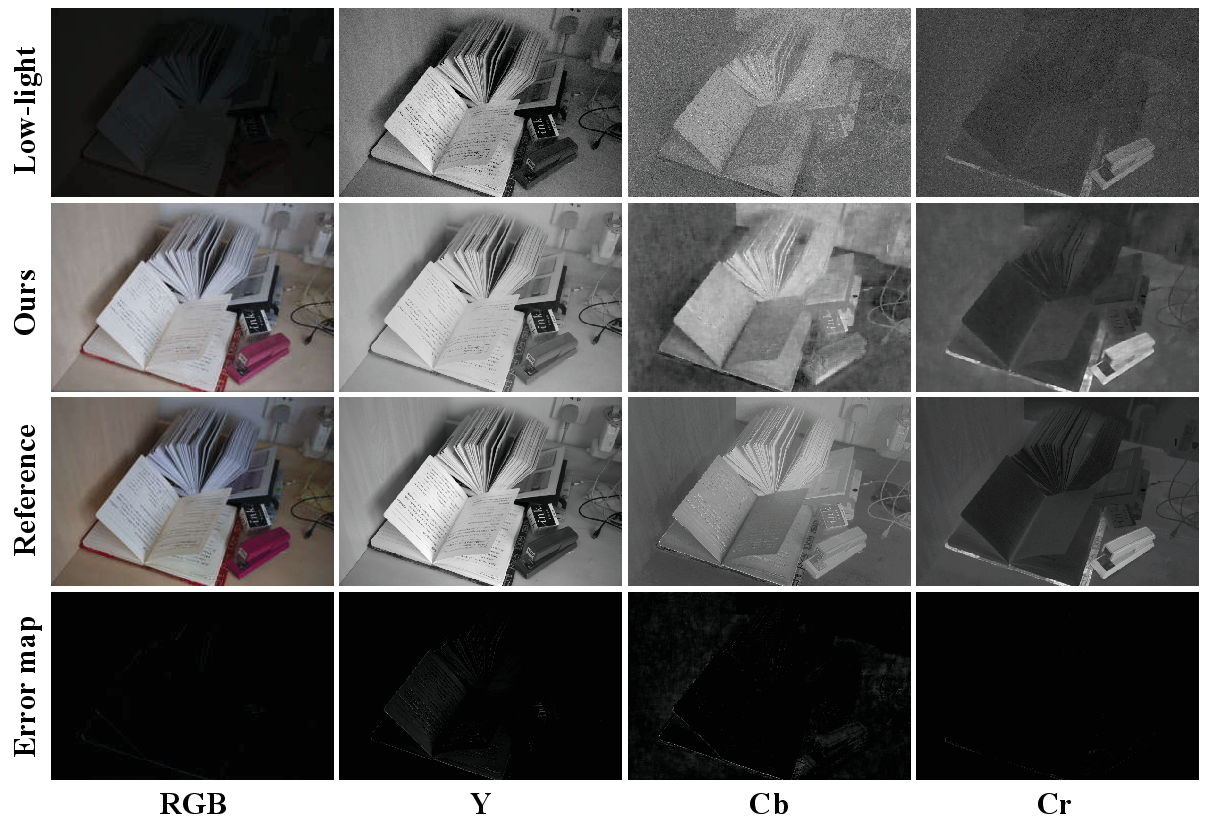}
\centering
\caption{The comparisons of before and after enhancement in different channels. The first row shows low-light images with their Y, Cb, and Cr channels. And the second row reveals the responding maps after enhancement. The third row is the responding maps of reference image. The last row presents the responding maps of error map. }
\end{figure}

\begin{table}[!t]
  \caption{The ablation study for LAN. The \textbf{bold} highlights the best results.}
\begin{center}
  \begin{tabular}{cccc}
  
    \toprule
    Methods   &	w/o Swin & w/o DACB & Ours  \\
    \midrule	 
 PSNR $\uparrow$ & 22.42	& 20.79    & \textbf{24.21}\\
SSIM $\uparrow$  & 0.8200	& 0.8118   & \textbf{0.8442}\\ 			
    \bottomrule
  \end{tabular}
\end{center}
\end{table}

\begin{table}[!t]
  \caption{The ablation study for Loss. The \textbf{bold} highlights the best results.}
\begin{center}
  \begin{tabular}{ccc}
  
    \toprule
    Methods   &	PSNR$\uparrow$ & SSIM$\uparrow$  \\
    \midrule	 
$\mathbb{L}_{LCDBNet}$              & 23.35	& 0.8315   \\
$\mathbb{L}_{LCDBNet}+ \mathbb{L}_{CRN}$   & 23.44  & 0.8334  \\ 
$\mathbb{L}_{LCDBNet}+ \mathbb{L}_{LAN}$   & 23.86	& 0.8398  \\ 
$\mathbb{L}_{LCDBNet}+ \mathbb{L}_{CRN} + \mathbb{L}_{LAN}$   & \textbf{24.21}	& \textbf{0.8442}\\ 		
    \bottomrule
  \end{tabular}
\end{center}
\end{table}

In LAN, we design a GLAB to simultaneously capture long-range information and local relation via a transformer channel and a convolution channel. To demonstrate their effectiveness, we separately remove each branch to conduct the same experiments. As shown in Table V, Swin branch and DACB branch gain 1.79 dB and 3.42 dB in PSNR, which demonstrates the significance of long-range and local information for luminance adjustment.

We propose a joint loss to end-to-end train our LCDBNet. To evaluate its effectiveness, we separately remove different sub-losses to conduct ablation experiments. The experimental results are shown in Table VI. One can observe that $\mathbb{L}_{CRN}$ can improve PSNR by 0.09dB and  $\mathbb{L}_{LAN}$ gains 0.51dB in terms of PSNR compared to $\mathbb{L}_{LCDBNet}$. The joint loss can achieve the best performance when two sub-losses, $\mathbb{L}_{CRN}$ and $\mathbb{L}_{LAN}$, are added into $\mathbb{L}_{LCDBNet}$. It demonstrates the effetiveness of our proposed joint loss.

\begin{figure}[!t]
\centering
\includegraphics[width=0.46\textwidth]{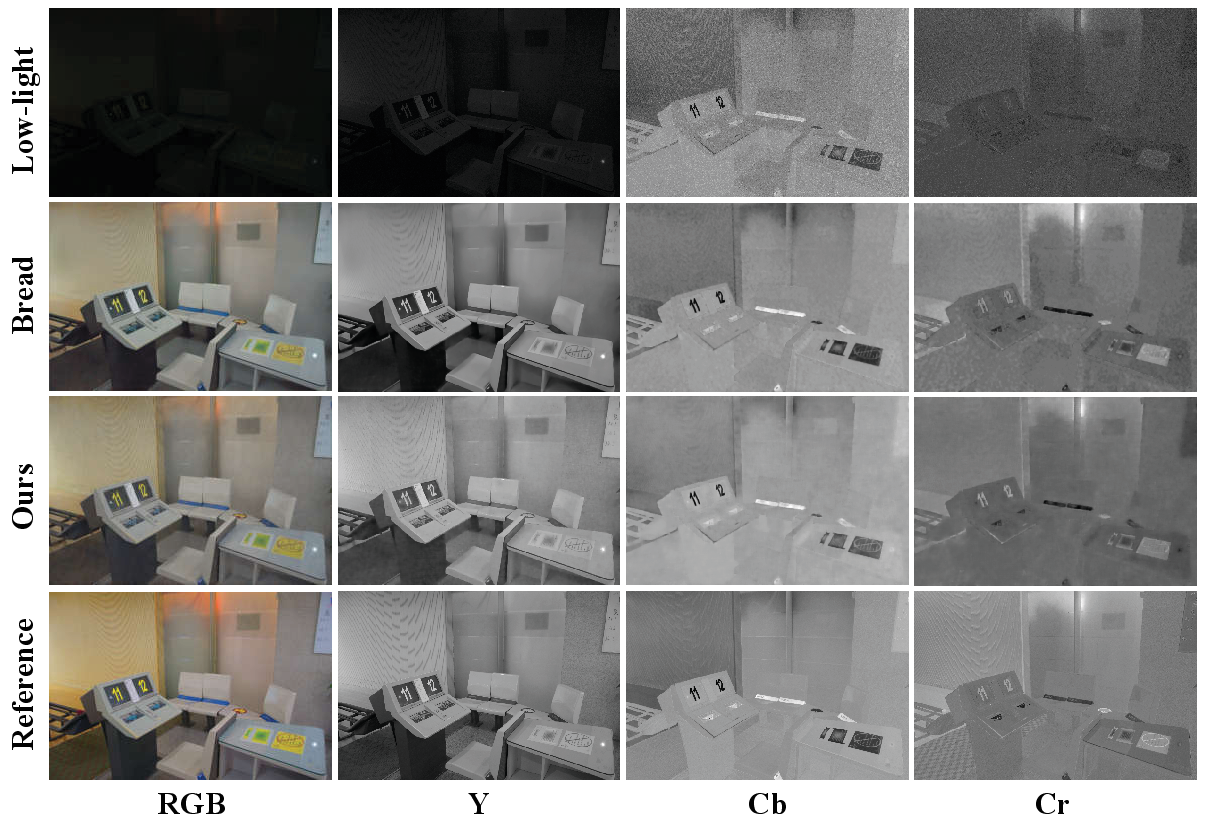}
\centering
\caption{The comparisons of Ours and Bread in different channels. The first row shows low-light images with their Y, Cb, and Cr channels. And the second row reveals the responding maps enhanced by Bread. The third row shows the enhancement images by Ours. The last row is the responding maps of reference image.}
\end{figure}

\begin{table}[!t] 
  \caption{The ablation study on different spaces. The \textbf{bold} highlights the best results. R-GB denotes that R channel is passed into LAN and GB channels are fed into CRN.}
\begin{center}
  \begin{tabular}{ccccc}
  
    \toprule
   Methods   &	R-GB	 & G-RB &B-GR  & Y-CbCr (Ours) \\
    \midrule	 
	PSNR $\uparrow$   & 	23.49    & 	22.98   & 	23.21  & 	\textbf{24.21}\\
	SSIM $\uparrow$   & 	0.8344  & 	0.8339  & 	0.8290 & 	\textbf{0.8442}\\	
	LPIPS $\downarrow$   & 	0.1402  & 0.1376  & 	0.1429 &  \textbf{0.1235}\\			
    \bottomrule
  \end{tabular}
\end{center}

\end{table}

\begin{table}[!t]
  \caption{The comparisons of running time and model parameters between ours and other models on LOL dataset.}
\begin{center}
  \begin{tabular}{cccc}
  
    \toprule
    Methods   &	Running time (s) & Parameter (M) & PSNR (dB) \\
    \midrule	 
LIME                 & 0.0783	& -           & 16.76\\
Zero-DCE         & 0.0047	& 0.079   & 16.77\\ 
SCI   &0.0010 &0.0003  &14.78\\
EnlightenGAN & 0.2278	& 8.64     & 17.48\\ 	
RetinexNet       & 0.5441	& 0.56     & 16.77\\ 	
KinD                & 0.7255	& 8.16	    & 20.38\\ 	
KinD++           & 8.8079	& 8.27    & 21.80\\ 	
MIRNet           & 1.2980	& 29.82	& 24.14\\ 	
URetinex         & 0.2157	& 0.40    & 21.33\\ 	
Bread              & 0.1208    & 3.80     & 22.96\\ 	
LLFormer       & 2.0680    & 24.52   & 23.65\\ 	
Ours               & 0.3524    & 7.36    & 24.21\\ 		
    \bottomrule
  \end{tabular}
\end{center}
\end{table}

\begin{figure*}[!t]
\centering
\includegraphics[width=0.88\textwidth]{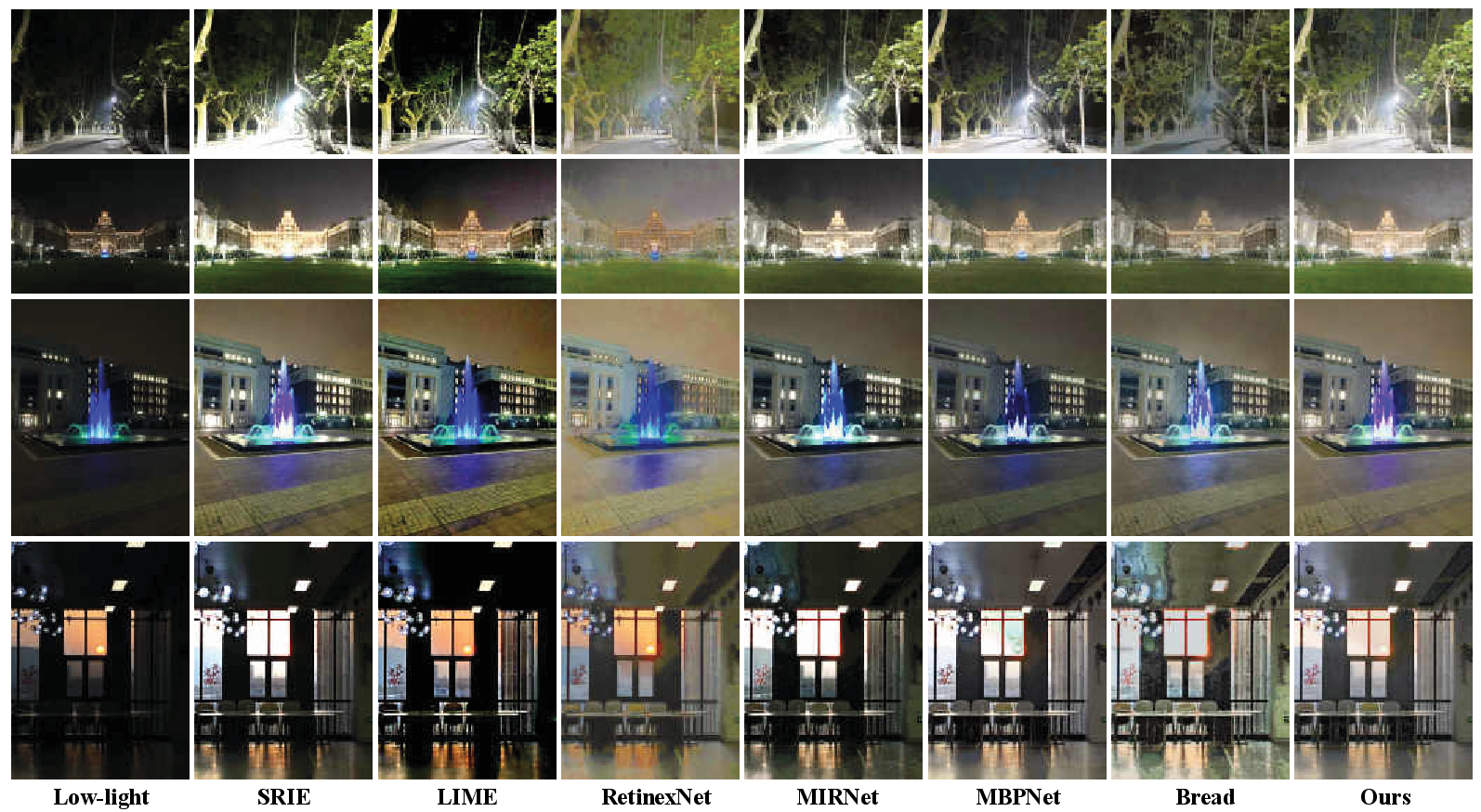}
\centering
\caption{The comparisons between ours and the representative low-light image enhancement methods on real-word low-light images.}
\end{figure*}

In order to justify our motivation, we compare the performance of LCDBNet on YCbCr and RGB spaces. Unlike low-light images on YCbCr space, low-light images on RGB space show the similar degradation degree and appearance in each channel. We respectively pass R channel, G channel, and B channel to LAN, and the remaining two channels into CRN, which are denoted as R-GB, G-RB, and B-RG. The corresponding results are reported in Table VII. As can be seen from quantitative indexes, YCbCr space holds significant advantages in PSNR, SSIM, and LPIPS compared to RGB space in low-light image enhancement. Thus, the above experiments substantiate the effectiveness of our design guidance.

Moreover, we explore the significance of our LCDBNet in YCbCr space. Y, Cb, and Cr maps before and after enhancement are demonstrated in Figure 14. The first row shows low-light images with their Y, Cb, and Cr channels. And the second rows reveal the responding maps after enhancement. The third row are the responding maps of reference image. The last row presents the responding maps of error map. As demonstrated in Figure 14, Y channel after enhancement is enlightened and chrominance maps (Cb and Cr) are obviously restored by removing the noise. We can see slight enhancement error from error maps. Zoom in for clearer review. It verifies our LCDBNet has significant advantage for low-light image enhancement in YCBCr space.

To demonstrate our advantages compared to other YCbCr-based enhancement methods, Figure 15 presents the comparisons between Ours and Bread in different channels. The first row shows low-light images with their Y, Cb, and Cr channels. And the second row reveals the responding maps enhanced by Bread. The third row shows the enhancement images by Ours. The last row is the responding maps of reference image. One can see that Ours shows more impressive enhancement results than Bread, and they substantiate the advantages of our proposed method.

\subsection{Running Time and Model Complexity}
To evaluate the efficiency of different models, we show the running time comparison and model parameter comparison. The test are performed on LOL dataset, the image size of which is $600 \times 400$. All results including running time, parameters, and PSNR are reported in Table VIII. The corresponding codes are downloaded from official codes and are tested with default parameters or pretrained models. One can see that our method has the best performance with relatively fast running speed and moderate parameters. Though MIRNet and LLFormer have comparative performance with ours in PSNR, their model parameters and test times are several times greater than ours. Zero-DCE and SCI ~\cite{67} show faster running time and are more lightweight models than ours, but their PSNRs are significantly lower than ours.

\subsection{Real-Word Low-Light Image Enhancement}
To evaluate the effectiveness of our model on real-word low-light image enhancement, we collected some real-word low-light images at night. Then, they are enhanced by our model and some representative low-light enhancement methods. Visual results are shown in Figure 16. As can be seen from that, LIME and SRIE cannot achieve desirable enhancements. RetinexNet generates unnatural enhanced results. MIRNet and MBPNet produce under-enhancement images. Bread yields obvious artifacts, especially in the fourth image. However, the processed images lightened by ours look more appealing than the images enhanced by other methods. Therefore, our method shows impressive potentials on real-word low-light image enhancement.

\section{Conclusion}
\label{sec:conclusion}
In this paper, we have presented a novel luminance and chrominance dual branch network (LCDBNet) for low-light image enhancement, which reformulates the problem of low-light image enhancement into two simple sub-tasks, namely, luminance adjustment and chrominance restoration. To tackle these tasks, luminance adjustment network (LAN) and chrominance restoration network (CRN) are designed to learn brightness-aware features and detail-sensitive representation, respectively. LAN inherits the advantages of convolution attention and transformer to model long-range and local pixel correlation, and CRN employs wavelet decomposition to extract high-frequency detail features. Then, we designed a fusion network (FN) to aggregate the learned features by LAN and CRN to yield the normal-light images. Extensive experiments on seven test sets demonstrate that our LCDBNet can recover normal-light images with vivid color and sharp texture. In future, we will explore our model to address other low-level vision tasks.
\bibliographystyle{IEEEtran}
\small
\bibliography{IEEEabrv,LCDBNet}

\end{document}